# Biometrics for Child Vaccination and Welfare: Persistence of Fingerprint Recognition for Infants and Toddlers

Anil K. Jain, Sunpreet S. Arora, Lacey Best-Rowden and Kai Cao
Department of Computer Science and Engineering
Michigan State University
East Lansing, Michigan 48824, United States
Email: {jain, arorasun, bestrow1, kaicao}@cse.msu.edu

Prem Sewak Sudhish
Department of Physics and Computer Science
Dayalbagh Educational Institute
Dayalbagh, Agra 282005, India
Email: pss@alumni.stanford.edu

Anjoo Bhatnagar
Pediatrician
Saran Ashram Hospital
Dayalbagh, Agra 282005, India
Email: dranjoo@gmail.com

## Abstract

*With a number of emerging applications requiring biometric recognition of children (e.g., tracking child vaccination schedules, identifying missing children and preventing newborn baby swaps in hospitals), investigating the temporal stability of biometric recognition accuracy for children is important. The persistence of recognition accuracy of three of the most commonly used biometric traits (fingerprints, face and iris) has been investigated for adults. However, persistence of biometric recognition accuracy has not been studied systematically for children in the age group of 0-4 years. Given that very young children are often uncooperative and do not comprehend or follow instructions, in our opinion, among all biometric modalities, fingerprints are the most viable for recognizing children. This is primarily because it is easier to capture fingerprints of young children compared to other biometric traits, e.g., iris, where a child needs to stare directly towards the camera to initiate iris capture. In this report, we detail our initiative to investigate the persistence of fingerprint recognition for children in the age group of 0-4 years. Based on preliminary results obtained for the data collected in the first phase of our study, use of fingerprints for recognition of 0-4 year-old children appears promising.*

## 1. Introduction

Biometric recognition has undoubtedly made great strides over the past century. The success of fingerprints in forensics and law enforcement has fueled a broad range of applications for biometric systems, ranging from national civil registries [8] to mobile devices [7]. The focal subject groups for biometric applications in law enforcement, government and personal devices are primarily adults and adolescents (e.g. typically over 5 years old in India's Aadhaar Program [8]). As a result, biometric vendors, system integrators and the research community have primarily focused on developing data capture and recognition solutions for adults. Furthermore, the pros and cons (e.g., in terms of recognition performance, cost, system vulnerability) of using different biometric traits (e.g., face, fingerprint, iris) for recognition of adults have been identified [15].

Fundamental premises about the use of biometrics are: (i) a biometric trait is unique to an individual, and (ii) its recognition performance does not change with time (persistence). While the uniqueness and persistence properties have been studied for the three most popular biometric traits (fingerprints [20], face [9] and iris [13]) for the adult population, there have been no studies reported for children (newborn to teens). Even for adults, theoretical results are based on relatively simple models, while empirical results are based on a limited sample of the population.

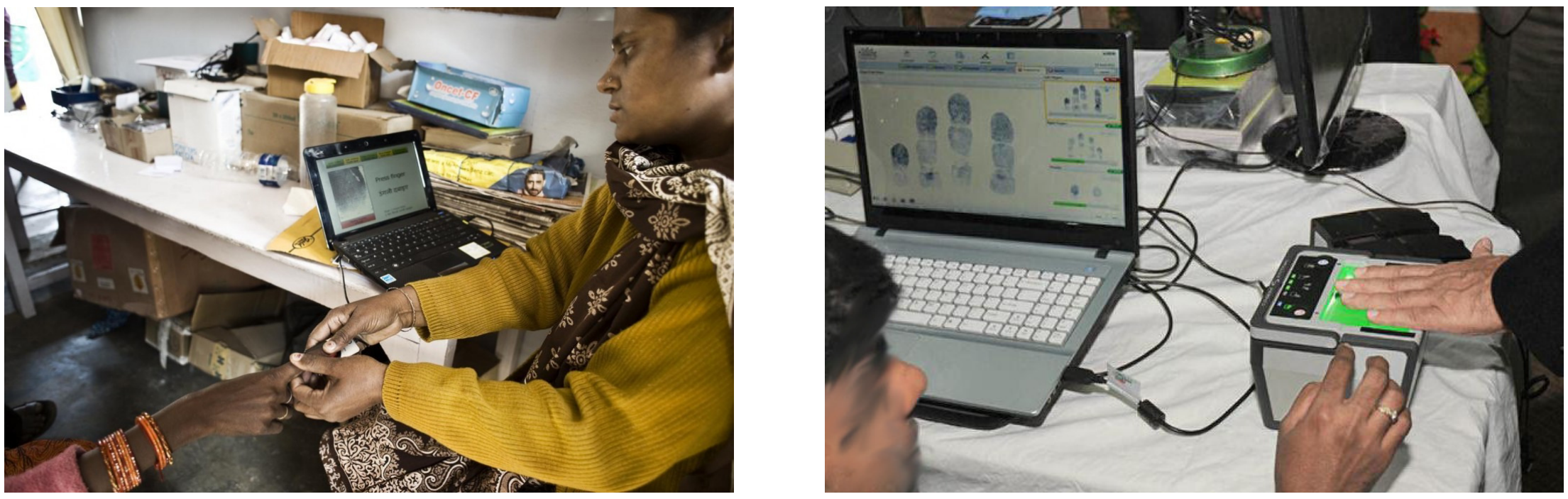

(a) (b)

*Figure 1: Potential applications requiring biometric recognition of children. (a) Operation ASHA's mobile healthcare e-compliance biometric system [5], and (b) Aadhaar civil registry project in India [8].*

With emerging biometric applications requiring recognition of younger subjects (0-4 years old), the research community as well as healthcare and government agencies are interested in answering the following question: can children[1] in the age range of 0-4 years be reliably recognized over time using their biometric traits? Examples of such applications include:

- *Tracking Child Vaccination Schedules and Welfare*: Every year, 25 million children (younger than 5 years old) do not receive the necessary vaccinations, and about 6.6 million children die due to vaccine preventable diseases [1]. Developing regions of Sub-Saharan Africa and South Asia together account for 4 out of every 5 of these deaths. Thus, a major goal of most governmental and non-governmental organizations is to eradicate the occurrence of vaccine preventable diseases by initiating nationwide routine and mandatory vaccination programs, especially in developing countries.

  Despite the existence of routine vaccination programs (e.g. Mission Indradhanush in India[2]), it is estimated that vaccine wastage rates are higher than 50% in some of the most challenging geographies[3]. Vaccinations are not being administered to the child in need due to the lack of an effective method to keep track of which child has been vaccinated and which vaccines have been administered to each child. To reduce vaccine wastage and increase vaccination coverage, it is, therefore, important to find a solution to track vaccination schedule of each child. Given that children, and even adults, in developing countries typically do not have any form of identification document, biometric identification offers a viable alternative to track vaccination schedules of each individual child.

  VaxTrac[1], a non-governmental organization (NGO) operating in the West African country of Benin and South Asian countries of Nepal and Bhutan, has developed a vaccine registry system based on fingerprints for tracking the vaccination schedules of children. Left and right thumb prints of the child and his mother are enrolled by a healthcare worker when the child is administered a vaccine for the first time. During subsequent visits to the health camp, the child's or his mother's fingerprints are used to determine whether the child has been administered any vaccines, and if so, which vaccines have been administered previously.

[1]We use the term child or children specifically for 0-4 year old infants and toddlers.
[2]http://www.nhp.gov.in/mission-indradhanush
[3]http://vaxtrac.com

While VaxTrac does not report the recognition accuracy for children's fingerprints, they mention that the accuracy is quite low; hence they simply use the mother's fingerprints instead[4]. Relying on the mother's biometric identity to recognize the child, however, is not a desirable solution for this application since the mother may not always accompany the child to the health camp or may have more than one child that requires vaccination.

Another NGO, Operation ASHA, in collaboration with Microsoft Research India, has developed an e-compliance biometric solution for tracking tuberculosis (TB) patients using fingerprints which is currently operational in 21 different locations in India [5] (Figure 1(a)). A social enterprise startup, SimPrints is developing a mobile (phone or tablet) based fingerprint scanner for associating an individual's fingerprints to their electronic health record registry[5]. To expand such innovative healthcare programs to cover children, it is necessary to investigate whether children can be recognized based on their fingerprints over time.

- *Identifying Missing Children*: Over 800,000 children go missing in the United States every year - one child almost every 40 seconds[6]. Many of them cannot be easily located and then identified. To alleviate this issue, a community service initiative called the National Child ID program was initiated in 1997 in the United States with the aim of fingerprinting 20 million children. The program's ID kit allows parents to capture and store their child's fingerprints. These fingerprints can later assist authorities in locating and identifying the child if the child goes missing. An important requirement for this application is to develop procedures to reliably match time-separated fingerprint impressions of children. The program has already grown into one of the largest identification efforts ever undertaken with over 26 million child ID kits distributed. However, no data is available on the performance of fingerprint recognition for this age group.

- *Preventing Newborn Baby Swaps*: Swapping of newborns after birth is a problem in developing countries because of inadequate facilities in maternity wards and overcrowding in hospitals[7]. To prevent newborn swapping, it is important to keep track of newborns and link their identities to their mothers. Although several hospitals use RFID bracelets for this purpose, the primary drawback of using such bracelets is that they are easily lost and exchanged. Consequently, the use of biometrics for tracking newborns has been advocated. Continual research efforts are needed to ascertain which biometric trait can be utilized for reliable identification of newborns.

- *Civil ID*: In 2009, the Aadhaar program was initiated by the Unique Identification Authority of India (UIDAI) with the aim of providing a unique ID (a 12-digit randomly generated number) to each of the 1.2 billion residents of India [8]. All 10 fingerprints and the 2 irises of a subject are captured and matched against the fingerprints and irises of all subjects enrolled in the system for de-duplication (Figure 1(b)). In current practice, biometrics are not captured for children below 5 years old, and although a unique 12 digit-ID is generated for them, the unique ID is linked to their parents. Biometrics are captured and the child is re-enrolled in the system when he turns 5 years old. Biometric data stored in the system is updated when the child turns 15 years old [3]. Given that about 29% of the Indian population is younger than 15 years of age [6], re-enrolling biometric data will require a massive effort both in terms of time and resources. This effort can be significantly reduced if minimum age can be reliably established at which child's biometric data can be captured, and be viable for identification based on biometric traits, thereby, reducing or even eliminating the need for re-enrollment.

[4]Based on our personal communication with VaxTrac.

[5]http://www.impatientoptimists.org/Posts/2013/10/Fingerprints-point-to-the-future-of-healthcare

[6]http://www.childidprogram.com/about-us

[7]http://timesofindia.indiatimes.com/city/ahmedabad/-Civil-Hospital-tags-newborns-to-prevent-baby-swapping/articleshow/6088759.cms?

| Biometric Trait | Required Degree of Subject Cooperation | Persistence | Parental Concerns |
|---|---|---|---|
| Face | Moderate (stare towards camera with neutral expression) | Low (facial aging) | Minor |
| Fingerprint | Moderate (allow the operator to hold the child's finger) | Potentially High | Moderate |
| Iris | High (open eyes and stare towards camera) | Potentially High | Major (IR illumination, obtrusive capture process) |
| Palmprint | High (open fist and allow operator to hold the palm and apply pressure) | Potentially High | Moderate |
| Footprint | High (removal of shoes and socks; allow operator to hold the foot and apply pressure) | Unknown | Minor (routinely used in U.S. hospitals for newborns) |

*Table 1: Comparison of the feasibility of using different biometric traits for recognizing children. The subjective entries in this table are solely based on the opinions of the authors.*

One of the primary difficulties in recognizing children based on their biometric traits is acquiring biometric data of sufficient quality for comparison and recognition. This is because younger subjects in the age group 0-4 may be uncooperative and not follow instructions. Among all the major biometric modalities proposed in the literature for recognizing children (fingerprints [12], face [10], iris [11], footprint [19] [16], and palmprint [19] [17]), we believe that fingerprints are the most viable for the following reasons:

- Capturing fingerprints requires relatively moderate cooperation from the child; the operator needs to hold the child's finger and place it on the fingerprint reader platen. This is more feasible in contrast to, for instance, capturing child's iris, where the subject needs to keep the eyes open and stare towards the camera, which is difficult if the child is asleep or uncooperative.
- Children's fingerprints are potentially highly persistent (a claim that we propose to investigate further in this study) in comparison to, for instance, facial characteristics which are known to change over time, particularly at young ages.
- Parents usually have minimal concerns in allowing fingerprint capture of their child once the procedure is demonstrated to them.
- It is possible to capture fingerprints of infants (age range of 0-2 years), who are not able to follow any instructions, while they are sitting comfortably in their mother's lap. In contrast, capturing footprints, for example, may require the subject to stand and place his foot properly on the reader.

Additionally, palmprints are difficult to capture because infants keep their fists closed (it is easier to capture a fingerprint by opening a finger), and parents may be concerned about iris cameras shining an IR illumination into the eye. In summary, based on a number of considerations such as the degree of subject cooperation needed, parental concerns, and our knowledge of the persistence of biometric traits, fingerprint, in our opinion, is the most feasible biometric trait for infant and toddler recognition (see Table 1).

In our earlier study, we had investigated capture and matching of fingerprints of infants and toddlers [14]. However, we were not able to systematically study the persistence property due to lack of temporal fingerprint data. In this study, our goal is to study the persistence of fingerprints of infants and toddlers. For this purpose, we have initiated a biometric data collection effort at the Saran Ashram Hospital, Dayalbagh, Agra, India in collaboration with Dayalbagh Educational Institute. In the first phase of data collection conducted in March 2015, we captured face and fingerprint images of left and right thumbs of 206 infants and toddlers. We describe this data collection effort in Section 2. Recognition performance of fingerprints and face for biometric identification of these children is reported in Sections 3.1 and 3.2, respectively. To study the persistence of fingerprint recognition for infants and toddlers, we plan to collect data from the same subjects at three additional time instances; September 2015 (Phase II) , December 2015 (Phase III), and March 2016 (Phase IV).

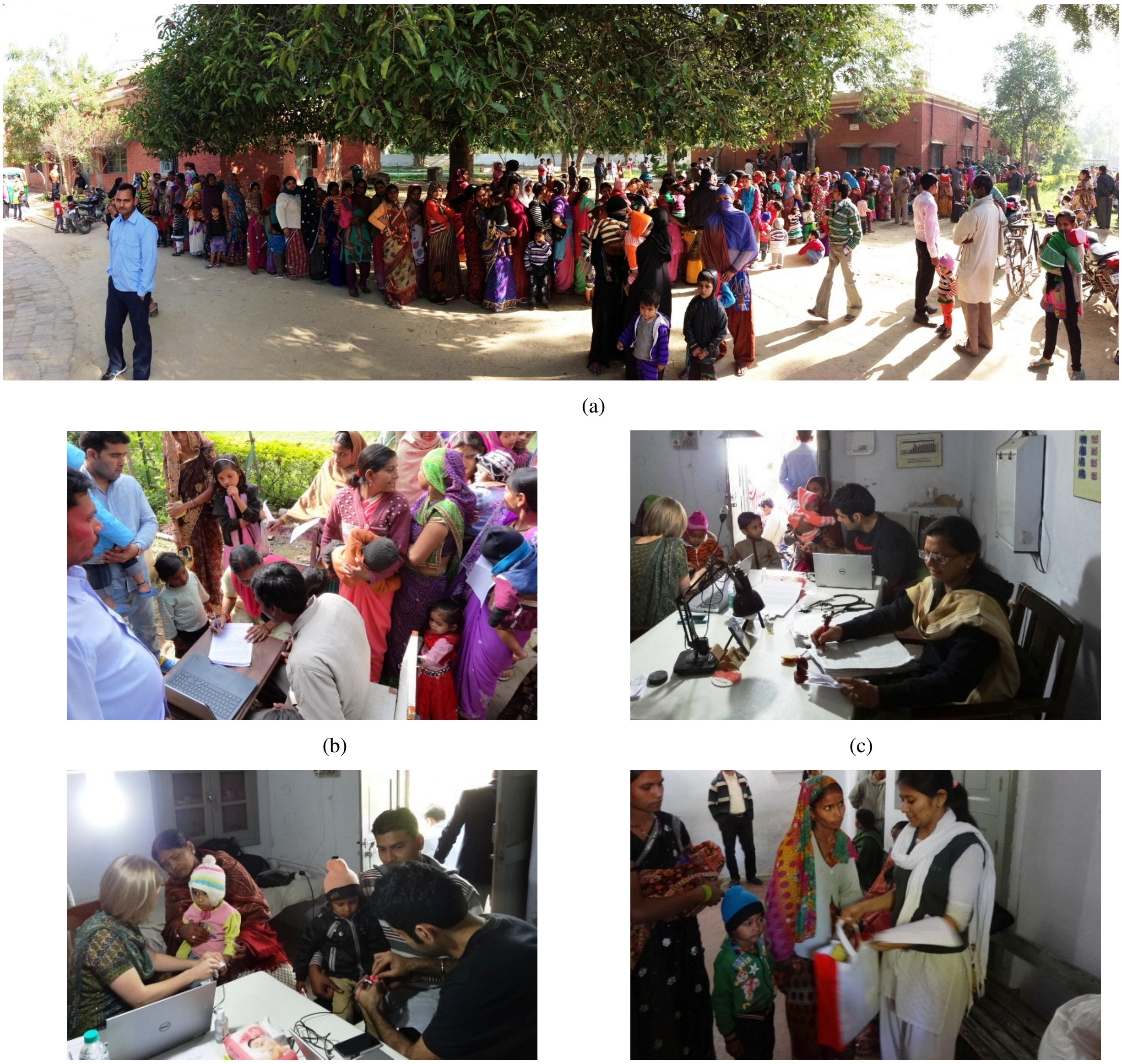

(a) (b) (c) (d) (e)

*Figure 2: Data collection effort at the Saran Ashram hospital in Dayalbagh, Agra, India. (a) Parents waiting outside the data collection room (Dr. Bhatnagar's office) to provide their child's biometric data, (b) parents signing the consent form agreeing to provide their child's biometric data, (c) two data collection stations in the doctor's office, (d) data capture at the two stations, and (e) handing out incentive package to the parents after data collection.*

## 2. Biometric Data Collection

Phase I of data collection took place over three days (March 8-10, 2015) at the Saran Ashram Hospital in Dayalbagh, Agra. Data was collected in a pediatrician's office (Dr. Bhatnagar), while the doctor was available monitoring the process and also examining other patients. Two different data collection stations were setup, each manned by a Ph.D. student, one doing research in fingerprint recognition and the other in face recognition. Each data collection station was equipped with a 500 ppi Digital Persona U.are.U 4500 HD fingerprint reader [18] (Figure 5) to collect fingerprint images of children. Face images were captured using the 8MP rear camera of iPhone 5/5s. Additionally, the child's name, age, gender, and address and contact number of the child's parents were collected to contact the parents for follow up data collection during our subsequent phases of data collection.

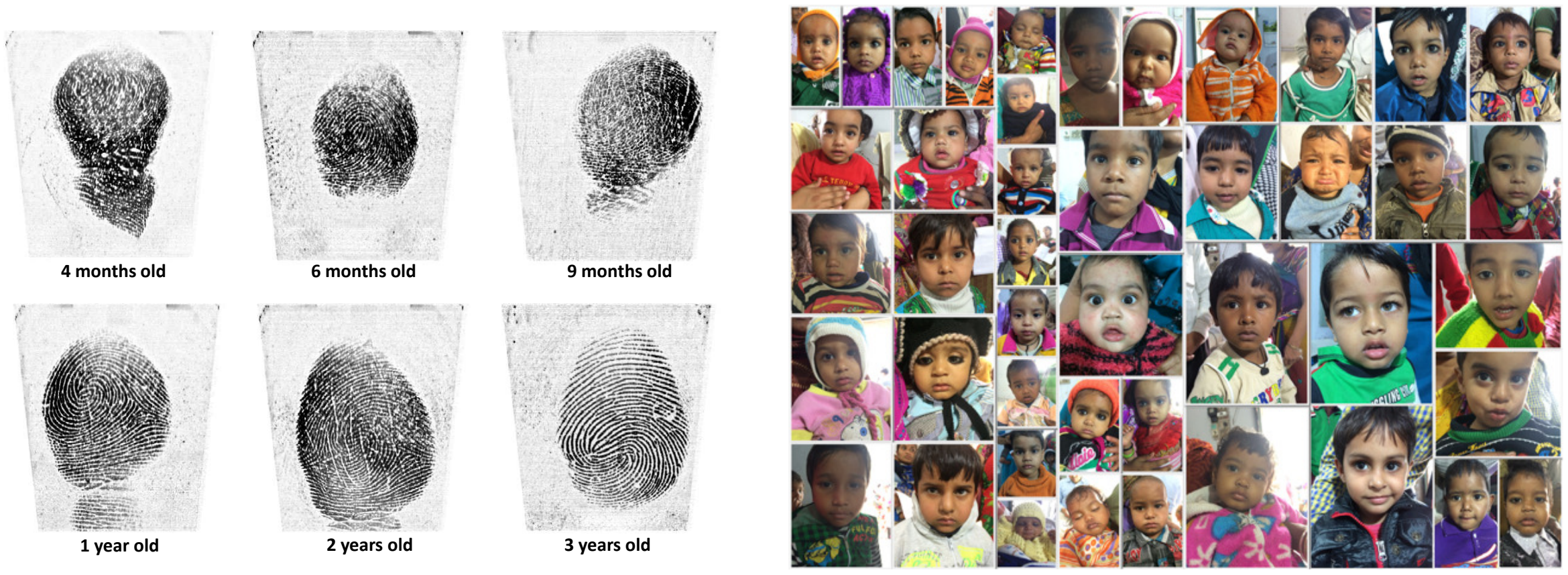

(a) (b)

*Figure 3: Sample fingerprint and face images from the database that was collected.*

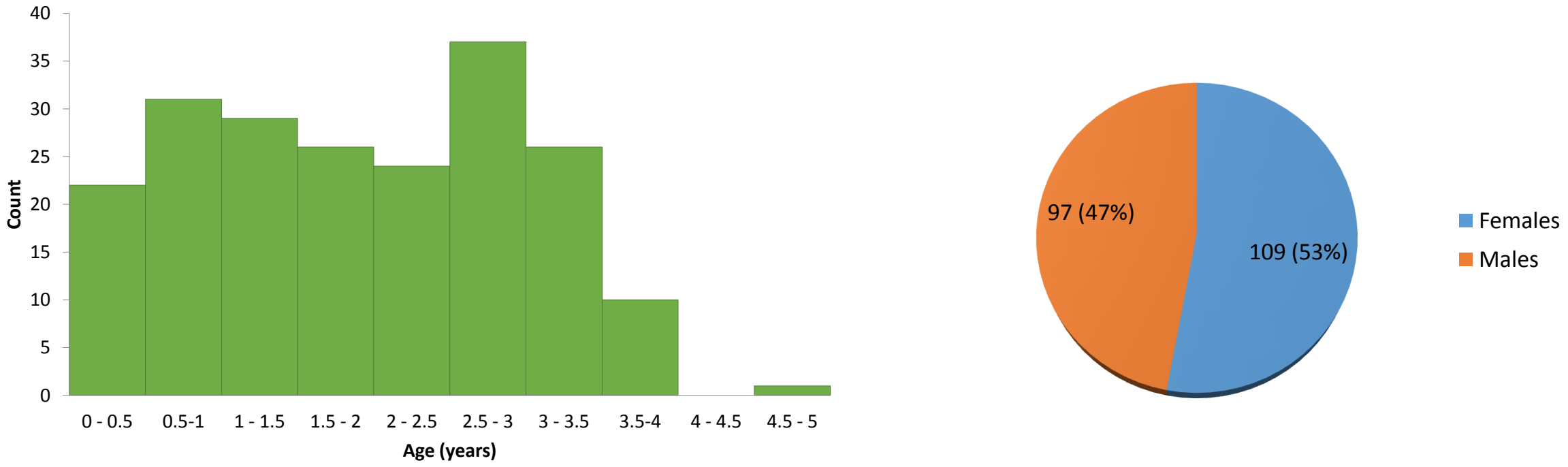

(a) (b)

*Figure 4: Age (a) and gender (b) distribution of the 206 subjects in the database.*

Figure 2 shows images of subjects, their parents and the data capture stations inside the pediatrician's office. Parents were required to sign a consent form (approved by the Ethics Committee of Dayalbagh Educational Institute, the hospital administration, as well as the MSU IRB Office) giving their consent to provide their child's fingerprint and face images. Face and fingerprint data were captured at one of the two data collection stations, and incentive (a bag consisting of rice, lentils, sugar and a toy for the child with a total value of about US$8) was handed out to the parents after data collection was complete.

## 2.1. Biometric Database

Initially, we had planned to capture three impressions each of the left thumb, left index finger, right thumb and right index finger of the child. However, considering that we had to maintain a high throughput (tens of subjects queued up to provide data even before we opened the doors of the data collection room), we decided to only capture three impressions each of the two thumbs. On day 1 of data collection, however, we could only capture two different impressions of the left and right thumbs of 36 subjects. This was because one of the data collection stations failed (software glitch with the fingerprint capture SDK). However, on days 2 and 3, we captured three impressions each of the two thumbs of 86 and 84 subjects, respectively. A total of 1,164 fingerprint images of 206 subjects were captured. Figure 3(a) shows sample fingerprint images of children of different ages. We also captured 3-5 face images per subject (810 face images from the 206 subjects) to investigate (i) whether face can be utilized as an additional cue to improve the overall recognition performance, and (ii) to display the face images of the top-K retrieved candidates in response to a fingerprint query so that the healthcare worker can ensure that the fingerprint match is indeed correct. Again, because of throughput requirements, face image capture was relatively unconstrained; illumination, pose and expression were not controlled during face capture. Figure 3(b) shows sample face images captured. The average time spent to capture fingerprint impressions of the two thumbs and face images was approximately 3 minutes per subject.

### 2.2. Demographic distribution

- *Gender*: Out of the 206 subjects who provided their biometric data, there were 97 males (about 47%) and 109 females (about 53%)
- *Age*: The age distribution of the subjects is shown in Figure 4(a). There are 10-37 subjects in each of the eight 6-month age brackets. (see Figure 4(b)).

### 2.3. Challenges and Observations in Data Collection

Below we summarize the key challenges faced, and some of our observations while capturing fingerprints and face images of children.

- *Dry fingers*: Due to relatively warm and dry environment in Agra, India in the month of March, the finger skin of many subjects was, noticeably, quite dry. Dry fingers would often not trigger the fingerprint reader automatically to capture prints. For these subjects, we used wet wipes to moisten the finger before capturing fingerprints.
- *Wet fingers*: A few of the younger subjects were sucking their thumb at the time of data capture. Because fingers which are wet result in poor fingerprint image quality, we dried their thumbs before capturing their prints.
- *Dirty fingers*: We observed that the hands of a few subjects were dirty. In such cases, we cleaned the subject's fingers with wet wipes. For one subject, we had to request the accompanying parent to wash the child's hands with soap and water. Once the hands were washed, we were able to capture the subject's prints.
- *Small finger size*: Although the Digital Persona U.are.U 4500 HD fingerprint reader is quite compact and ergonomically well designed (see Figure 5), for younger subjects (less than 6 months old) with very small fingers, we observed that placing the finger properly on the reader platen was challenging. As a result, only a partial fingerprint could be acquired in many such cases. Small fingers also sometimes did not trigger the reader to automatically capture a fingerprint. In such cases, repeated attempts had to be made to capture the prints successfully.
- *Manual vs. auto capture:* We initially tried to capture fingerprints of a few children using a 1000 ppi fingerprint reader (NEC PU900-10 [4]) in manual capture mode. It was challenging to capture fingerprints in the manual mode because holding the child's hand steady on the reader platen and manually triggering the capture at the same time was a challenge. We also observed slight motion blur in manually captured fingerprint images. In our experience, automatic capture is more convenient and efficient for capturing fingerprints of young children. We expect to use this 1000 ppi fingerprint reader in auto-capture mode during subsequent phases of data collection at Agra.

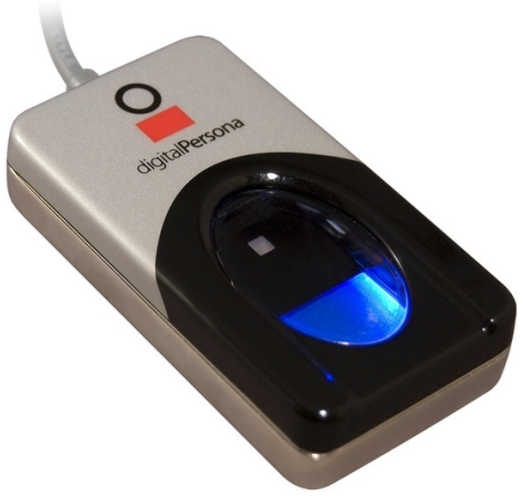

(a)

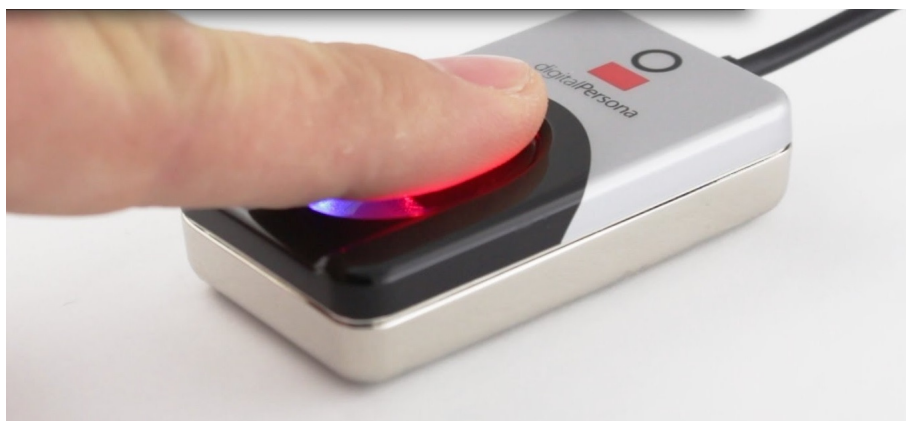

(b)

*Figure 5: (a) The 500 ppi Digital Persona U.are.U 4500 HD fingerprint reader [18] used to collect child fingerprints. The approximate reader size is 65 mm x 36 mm x 15.56 mm and it weighs 105 grams. (b) A red light glows once the finger is placed on the reader platen to trigger the fingerprint image capture.*

- *The "fear" of vaccination*: Since we were capturing data in a doctor's office, some subjects were uncomfortable (started crying) because they thought they were being administered vaccination. Capturing both fingerprints and face was difficult for such subjects. However, many children calmed down and became interested in the process after being attracted to the glowing blue and red lights of the fingerprint reader.

## 3. Experiments

Fingerprint and face matching performances were evaluated in both the verification (1:1 comparison) and identification (1:N comparison) modes of operation using fingerprint and face commercial-off-the-shelf (COTS) matchers. Two different metrics are used to evaluate verification performance, (i) *true accept rate (TAR)*, i.e. how many children, amongst those previously enrolled, can be successfully verified, and (ii) *false accept rate (FAR)*, i.e. how many children, amongst those not previously enrolled, are incorrectly determined to have been previously enrolled. Although it is ideally desirable to maximize true accepts while minimizing false accepts, in reality, there is a trade-off between TAR and FAR. Receiver Operating Characteristic (ROC) curve is plotted as TAR v. FAR at different operating thresholds to indicate the verification performance.

For identification, typically, a candidate list of the top-K matches is retrieved, and the retrieval rank of the true mate in the candidate list is used as an evaluation metric. Cumulative Match Characteristic (CMC) curve, where each point on the curve denotes whether the true mate was retrieved at rank $\leq i$ in the candidate list, is plotted to indicate the identification performance. Currently, the experiments are reported only for "closed set" identification scenario where the query subject is assumed to be present in the database. Below, we describe the verification and identification experiments we conducted.

### 3.1. Matching Fingerprint Images

We establish the baseline recognition performance by conducting experiments using two different COTS SDKs, a tenprint SDK (COTS-T) and a latent fingerprint SDK (COTS-L). All fingerprint images are upsampled by a factor of 1.8 using MATLAB's *imresize* function before inputting them to the SDKs. The upscaling is necessary to ensure that the ridge spacing of infant's and toddler's fingerprint images (around 4-5 pixels) closely approximates that of adults (around 8-9 pixels), in turn, facilitating feature extraction using the SDKs (see Figure 6).

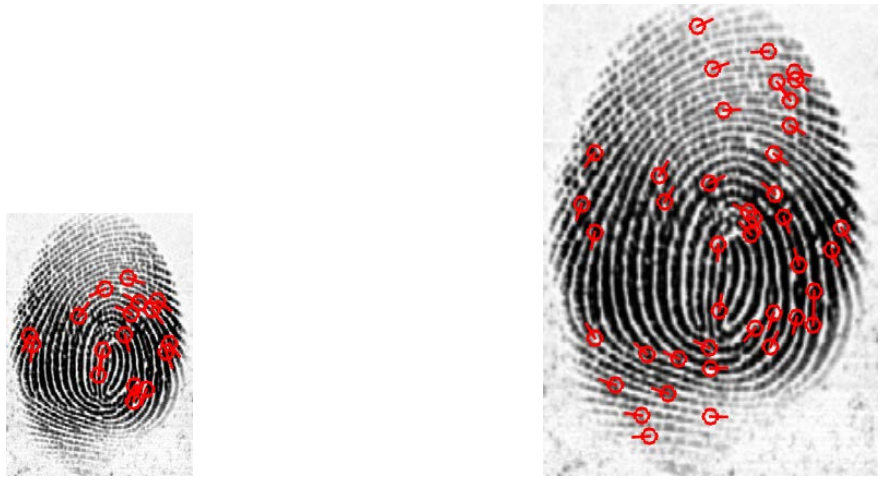

(a) (b)

*Figure 6: Extracted minutiae marked on (a) the original fingerprint image (17 minutiae were extracted), and (b) the upsampled fingerprint image (51 minutiae were extracted) of a 7 months-old child using the tenprint SDK (COTS-T).*

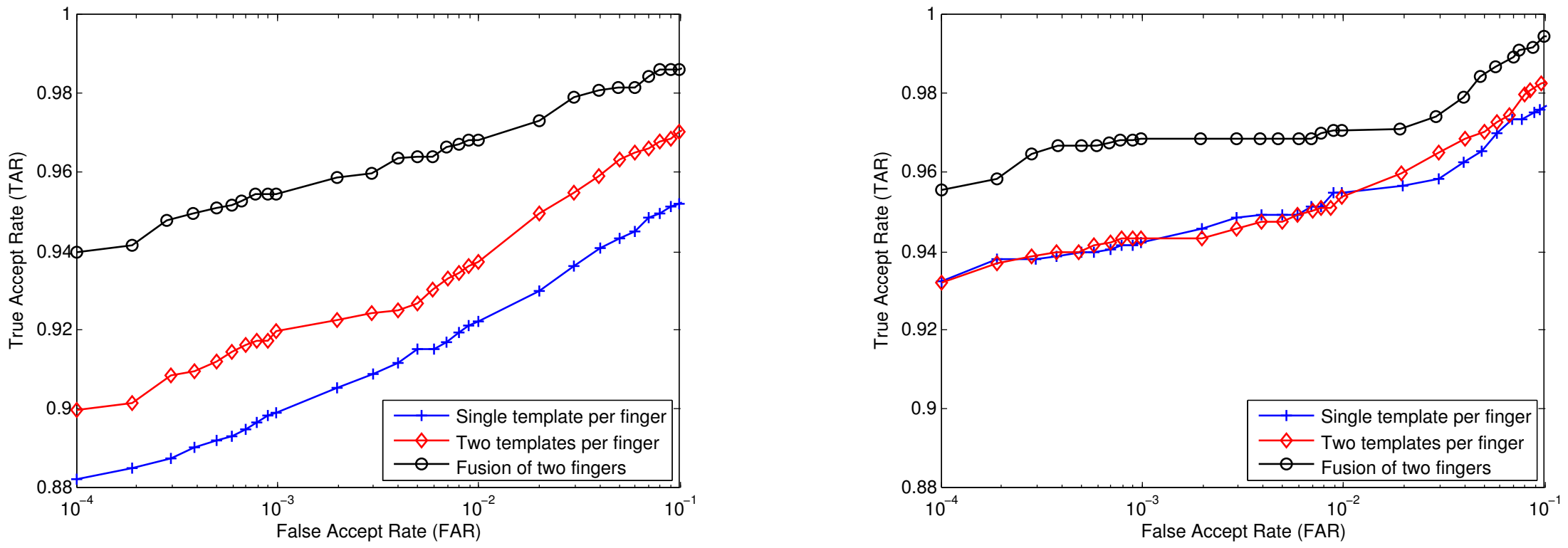

(a) COTS-T (b) COTS-L

*Figure 7: Receiver Operating Characteristics (ROC) curves for fingerprint verification experiments conducted using the two fingerprint SDKs, COTS-T and COTS-L.*

### 3.1.1 Fingerprint Verification

The verification protocol used in our experiments is analogous to that used in the Fingerprint Verification Competition (FVC) [2]. The following verification scenarios are considered:

1. *Single template per finger*: Assume one of the impressions of a finger (among the two or three impressions per finger acquired) is enrolled in the database. The remaining impressions of the finger are matched against the enrolled template to generate the genuine score distribution. The impostor score distribution is generated by comparing the first acquired impressions of two different fingers. The total number of genuine and impostor comparisons are 1,128 and 84,666, respectively. For this experiment, TARs of 89.92% and 94.24%, are obtained at FAR of 0.1% using COTS-T and COTS-L, respectively.

2. *Two templates per finger*: Now assume that two templates are enrolled for each finger. A probe fingerprint query is matched against the two templates of an enrolled finger, and the two comparison scores are fused to obtain a single score. Each impression of a finger is matched against the remaining two impressions of the same finger to compute the genuine score distribution. The first impression of each finger is matched against the first and second impressions of the other fingers to generate the impostor score distribution. The total number of genuine and impostor comparisons are 1,146 and 84,666, respectively. For this scenario, TARs of 91.99% and 94.33% are obtained by average fusion of scores from the two templates at a FAR of 0.1% using COTS-T and COTS-L, respectively. Note that since we only captured two fingerprint impressions on day 1, the accuracy improvement reported here is only for day 2 and day 3 subjects. Improvement in TAR by using two templates per finger over one template per finger is minimal.

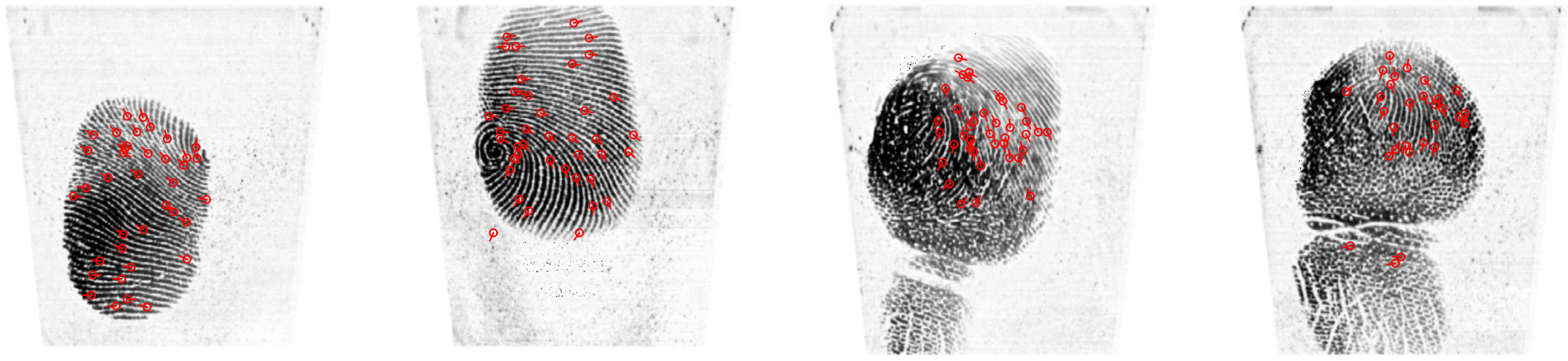

(a) (b) (c) (d)

*Figure 8: Examples of false rejects due to small overlap and large distortion between fingerprint impressions (a) and (b) of a 2 year old child, and poor quality fingerprint impressions (c) and (d) of a 4 month old child.*

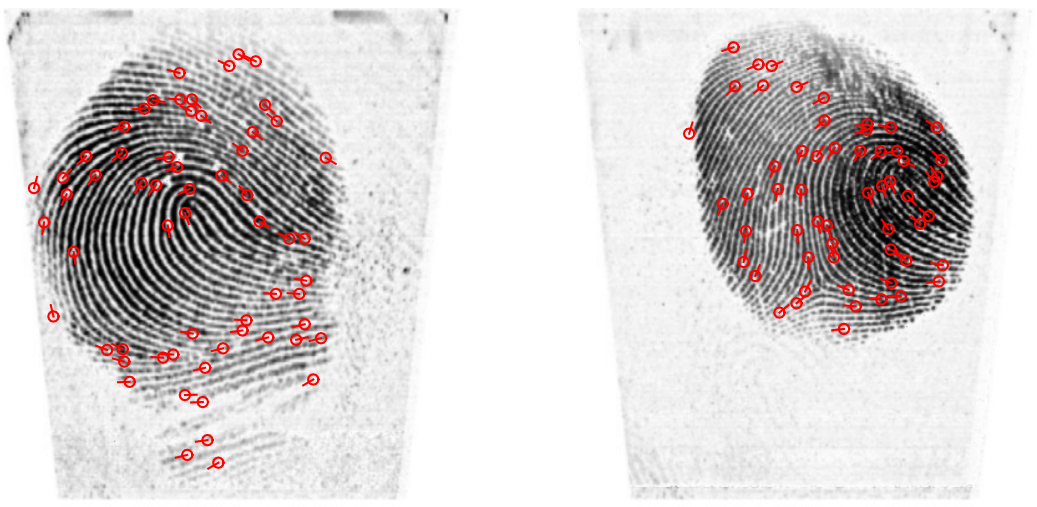

(a) (b)

*Figure 9: Example of a false accept due to similar ridge structure in fingerprint images (a) and (b) of two different subjects.*

3. *Fusion of two fingers*: Now, two fingerprints, one each from the two enrolled fingers, are compared against one (for day 1 subjects) or two (for day 2 and day 3 subjects) templates of the respective fingers for verification. Average fusion of the comparison scores is used as the final score. Genuine and impostor score distributions are computed analogous to the two enrolled template scenario. The total number of genuine and impostor comparisons 3,330 and 42,230, respectively. Fusion of scores obtained by comparison against enrolled templates from two fingers, improves the TAR at a FAR of 0.1% to 95.46% and 96.88% for COTS-T and COTS-L, respectively. Using two fingers for verification is significantly better than using a single finger.

ROC curves for verification experiments are shown in Figure 7. Figure 8 shows two false reject examples, and Figure 9 shows a false accept example. Fusion of two fingers results in the best verification performance. Furthermore, the following two experiments are conducted to investigate the effect of age and gender on matching performance. All results reported for these experiments are based on two finger fusion using one or two templates per finger.

- *Effect of age*: All the subjects in our database are divided into three groups, (i) 0-6 months old, (ii) 6-12 months old and (iii) over 12 months old. For each group, fingerprints of subjects within the group are used for computing the genuine score whereas impostor score distribution is obtained by fingerprint comparisons across all age groups. The ROC curves and distributions of the number of minutiae for the three age groups are shown in Figures 10 and 12, respectively. The conclusions derived from this experiment are as follows: (i) Recognition performance is significantly lower for subjects under 6 months old (69.19% and 73.74%, respectively, for COTS-T and COTS-L) compared to older subjects in our database. This can be attributed primarily to the generally poor quality of (partial) fingerprints of subjects younger than 6 months. (ii) Recognition performance is stable for subjects older than 12 months (99.74% and 100%, respectively, using COTS-T and COTS-L). (iii) The number of minutiae extracted by the SDK increases with the increase in the subject age. This is mainly because a larger finger area is captured for subjects older than 12 months compared to subjects below 6 months of age.

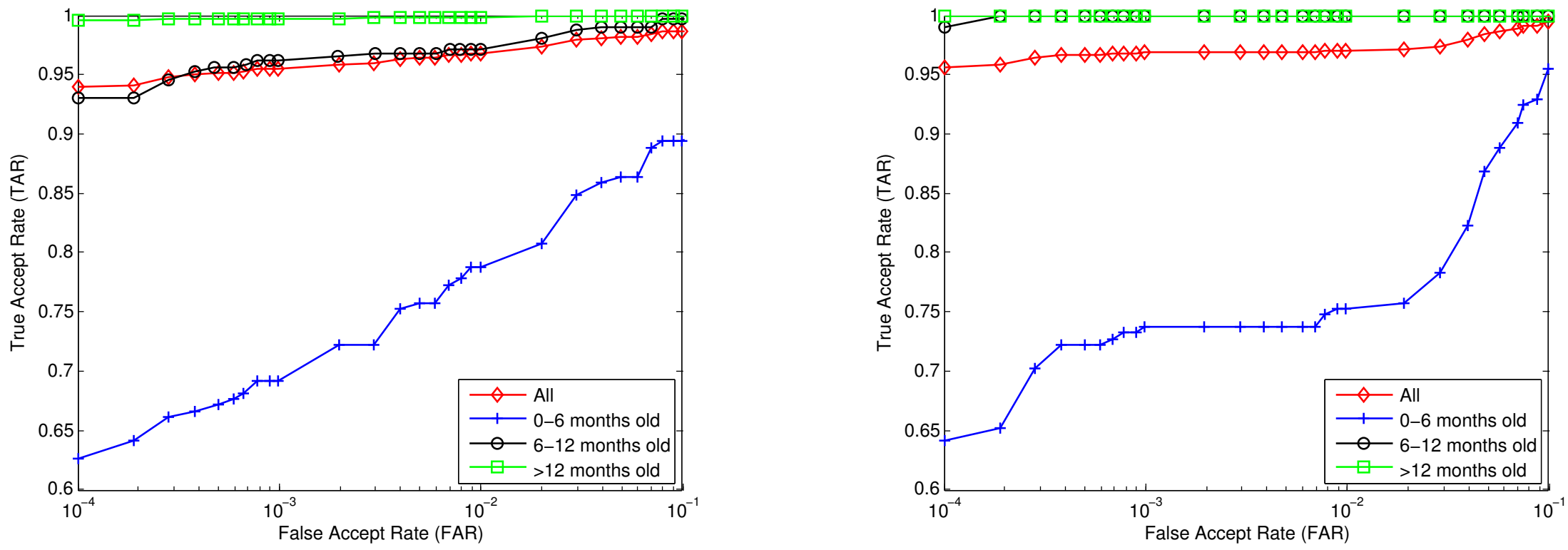

(a) COTS-T (b) COTS-L

*Figure 10: Receiver Operating Characteristics (ROC) curves for different age groups for fingerprint verification experiments conducted using the two fingerprint SDKs, COTS-T and COTS-L.*

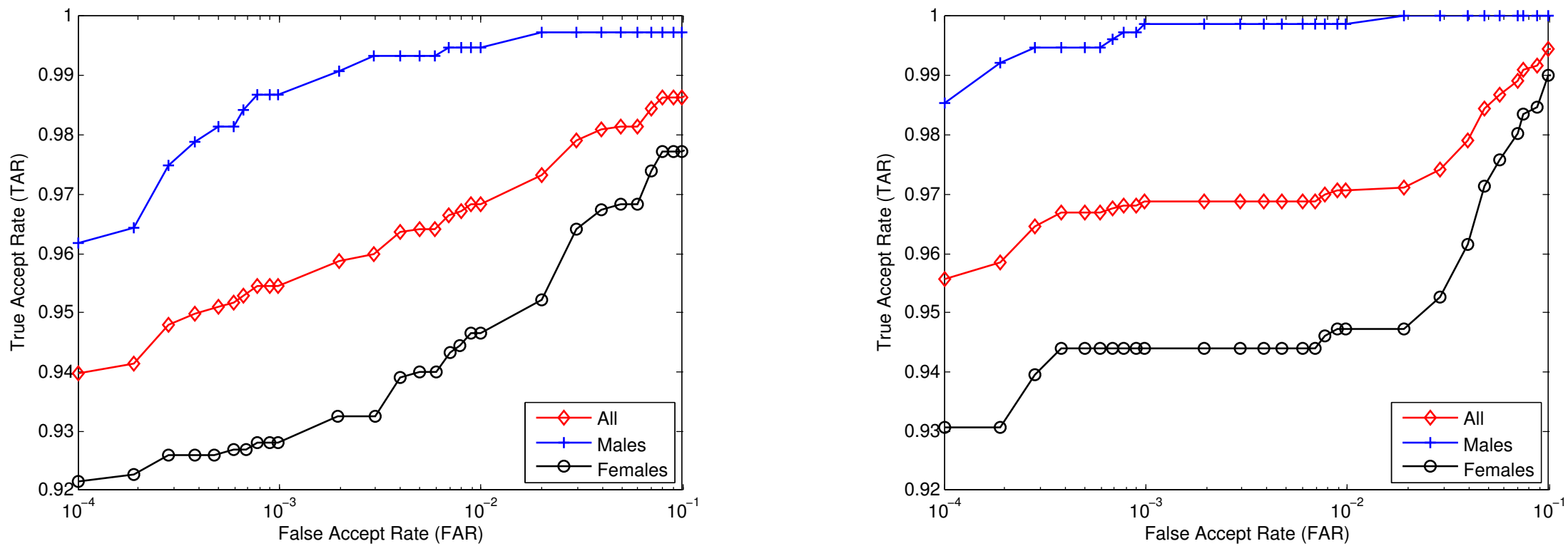

(a) COTS-T (b) COTS-L

*Figure 11: Receiver Operating Characteristics (ROC) curves for different genders for fingerprint verification experiments conducted using the two fingerprint SDKs, COTS-T and COTS-L.*

- *Effect of gender*: Figure 11 shows the ROC curves for (i) all subjects, (ii) only female, and (iii) only male subjects. Recognition performance of male subjects is significantly higher than that of females, even though there are 12 male subjects under 6 months old in comparison to 10 female subjects in the same age group. In our opinion, a key reason for this difference in the recognition performance is that male subjects have relatively larger fingers and ridge spacings on their fingers than female subjects. Hence, it is easier to capture good quality fingerprints for males compared to females in our target population.

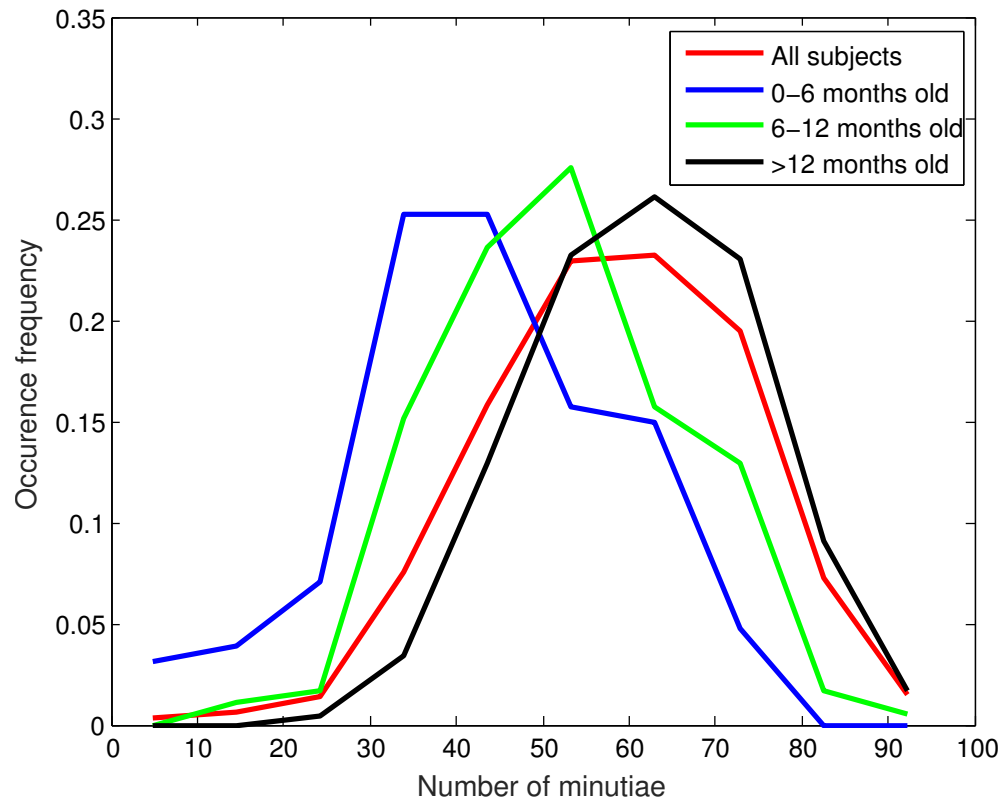

*Figure 12: Distribution of the extracted minutiae from fingerprint images (upsampled) of subjects belonging to different age groups using the COTS tenprint SDK (COTS-T). Note that as child grows older, more minutia points can be extracted, resulting in a higher recognition accuracy.*

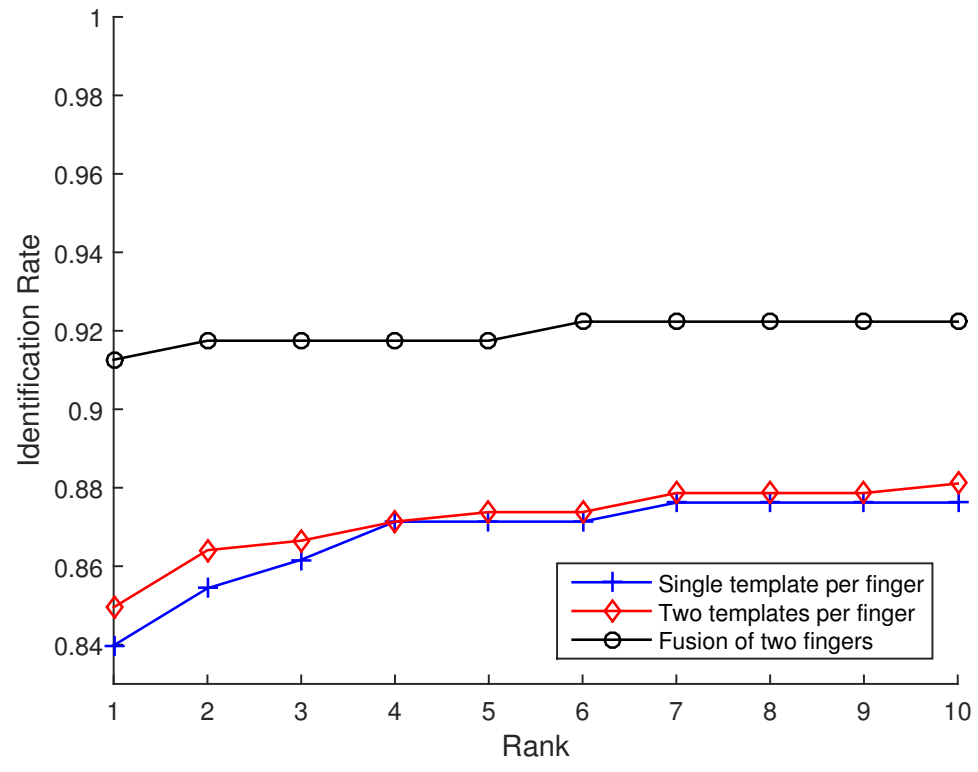

(a) COTS-T

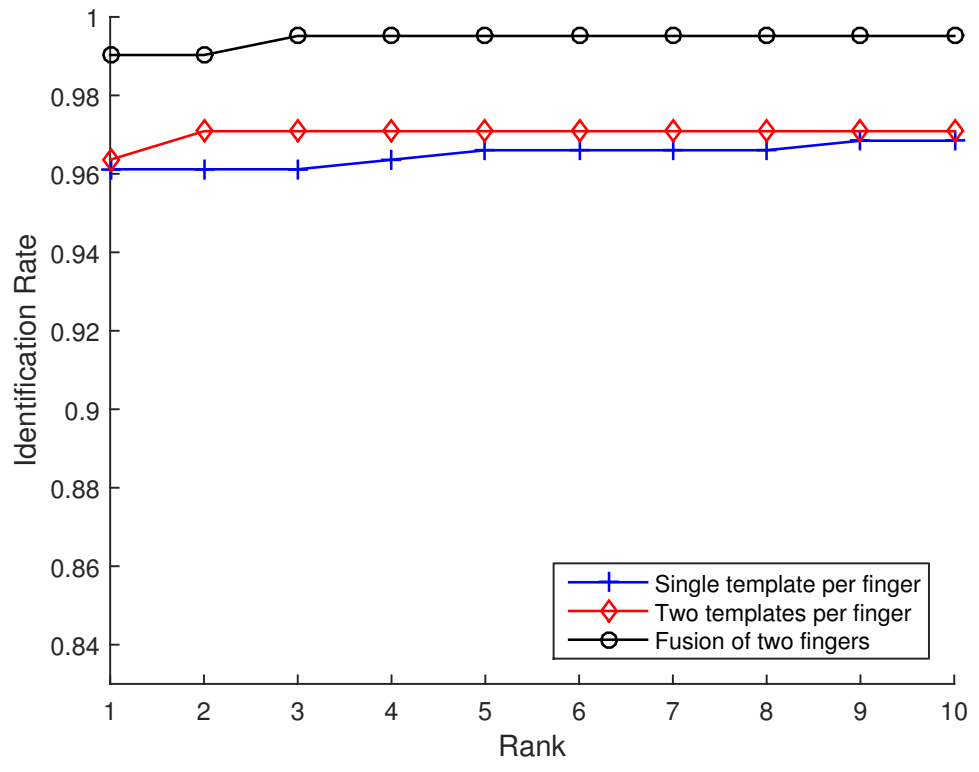

(b) COTS-L

*Figure 13: Cumulative Match Characteristics (CMC) curves for fingerprint identification experiments conducted using the two fingerprint SDKs, COTS-T and COTS-L.*

### 3.1.2 Identification Experiments

Analogous to verification, three identification experiments are conducted, matching against (i) single enrolled template of a finger, (ii) two enrolled templates of a finger, and (iii) two different enrolled fingers. An additional 32,768 fingerprints of 16,384 children (one impression each of the left and right thumb), collected by VaxTrac[8], are used to extend the size of the gallery for these experiments.

[8]http://www.vaxtrac.com

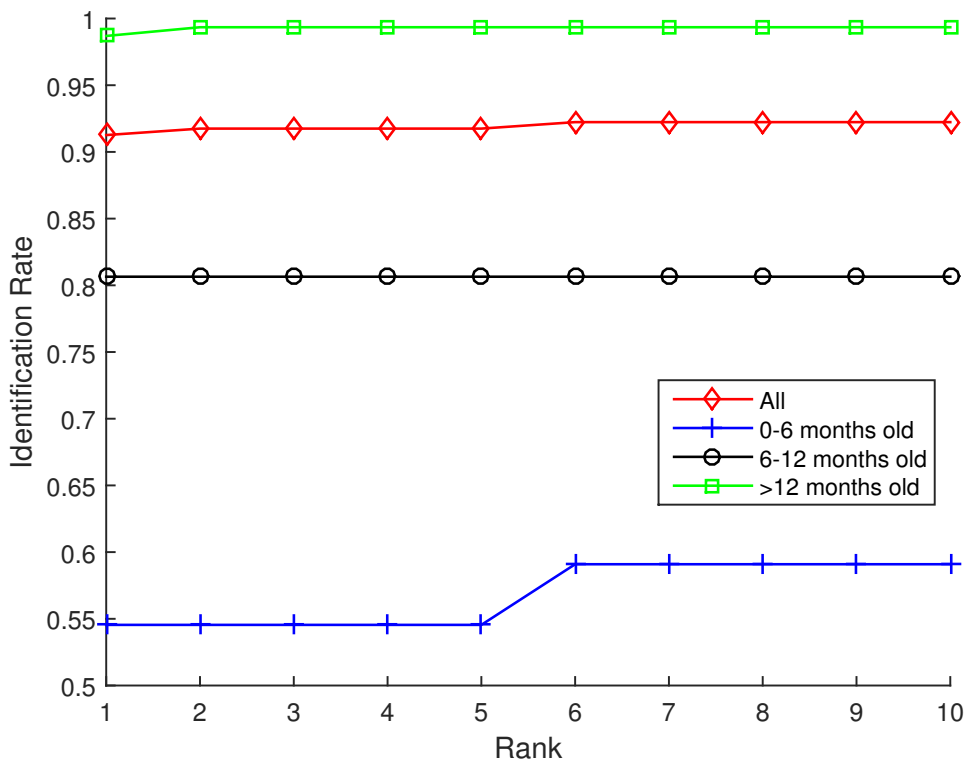

(a) COTS-T

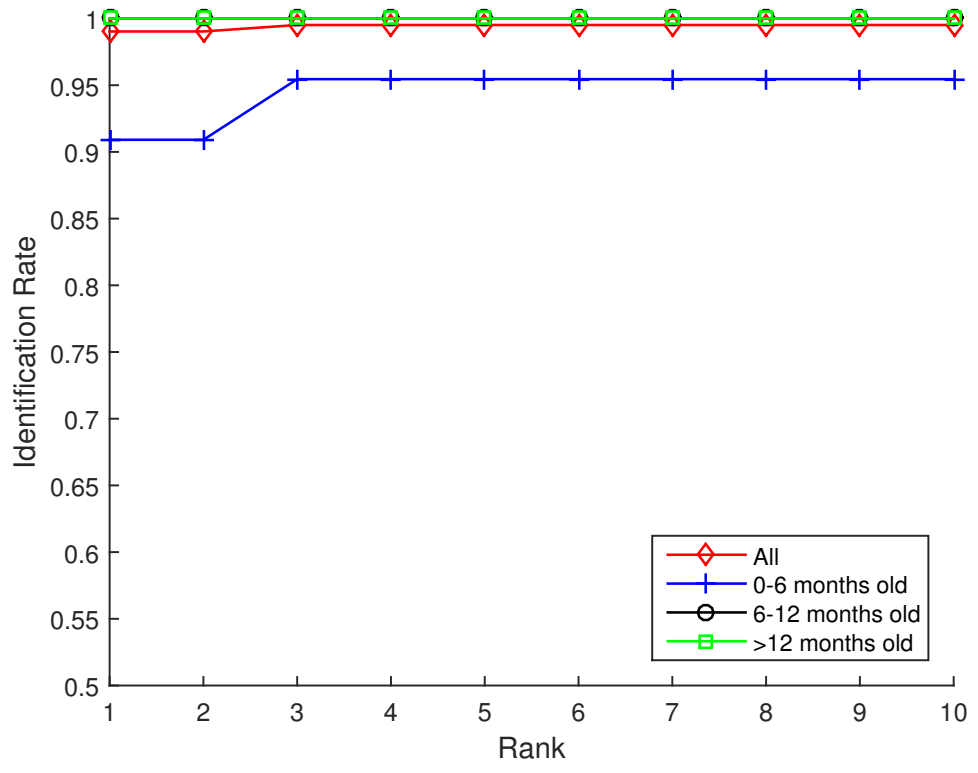

(b) COTS-L

*Figure 14: Cumulative Match Characteristics (CMC) curves for different age groups for fingerprint identification experiments conducted using COTS-T and COTS-L.*

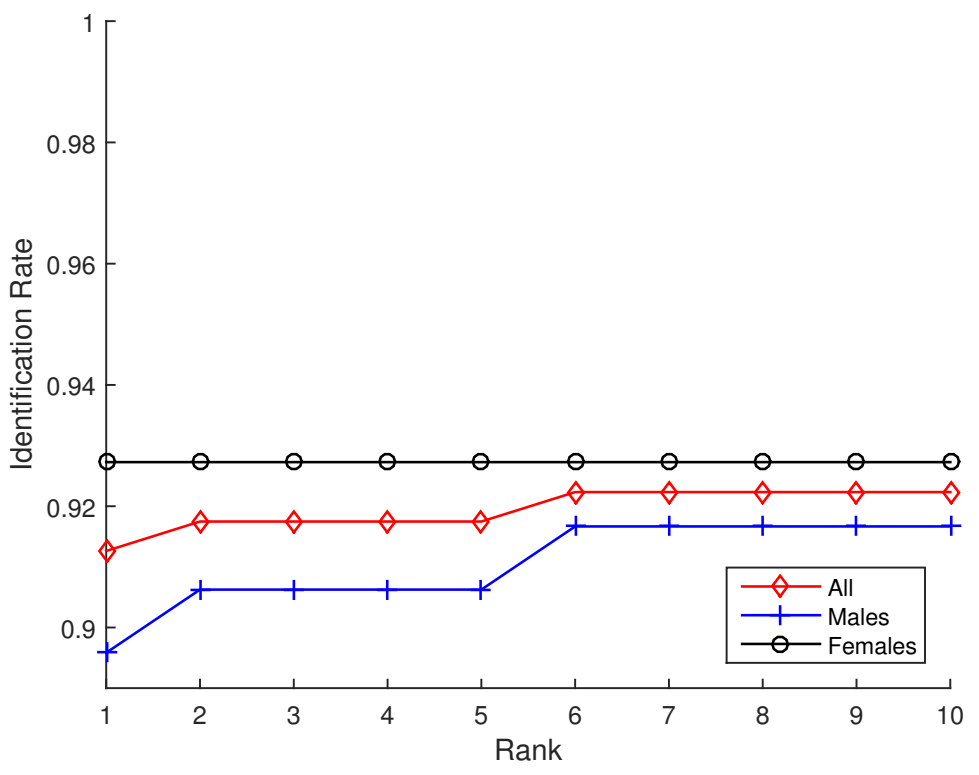

(a) COTS-T

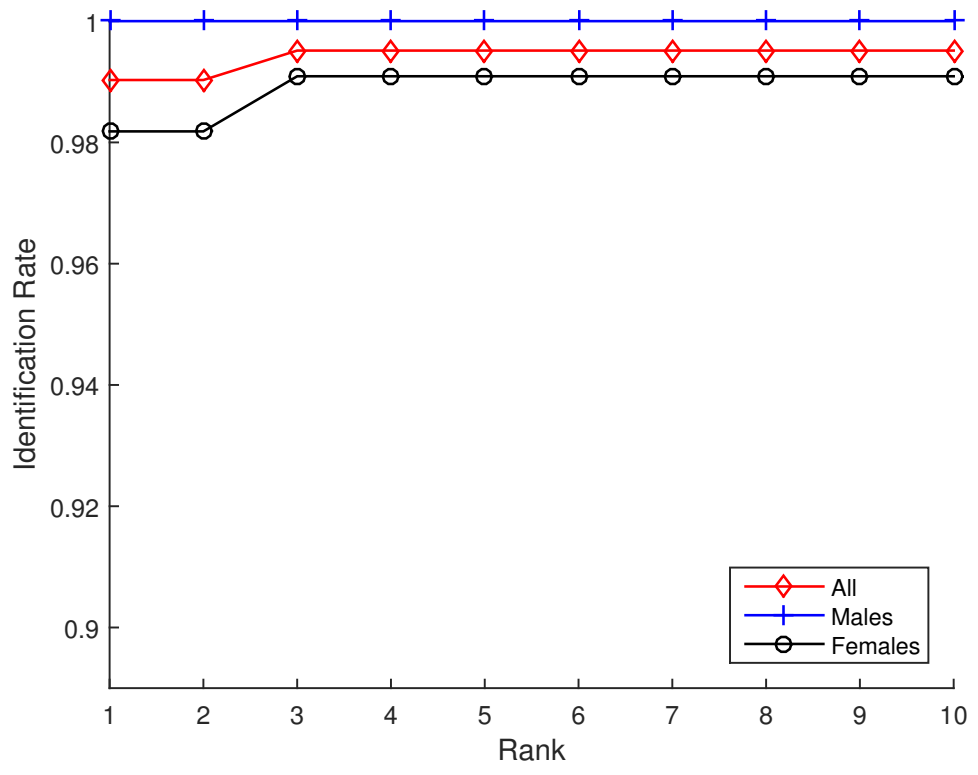

(b) COTS-L

*Figure 15: Cumulative Match Characteristics (CMC) curves for different genders for fingerprint identification experiments conducted using COTS-T and COTS-L.*

In experiment (i), the second impression of each finger is used in the gallery while the first one is used as a query. For experiments (ii) and (iii), the second and the third impressions are used in the gallery, and the first one is used as a query analogous to experiment (i). The rank-1 identification accuracy of matching against a single template is 83.98% and 96.12%, respectively, for COTS-A and COTS-B. Fusion of results from matching against two templates improves the rank-1 identification accuracy of COTS-A to 84.95% and of COTS-B to 96.36%. The rank-1 accuracy further improves to 91.26% and 99.03% for COTS-A and COTS-B, respectively, by fusion of two different fingers. The CMC curves for the three experiments are shown in Figure 13.

The identification accuracies for different age groups and gender are also investigated. For different age groups, trends similar to those observed in verification experiments are observed (Figure 14). In case of different gender, identification accuracies for male and female subjects are comparable (Figure 15).

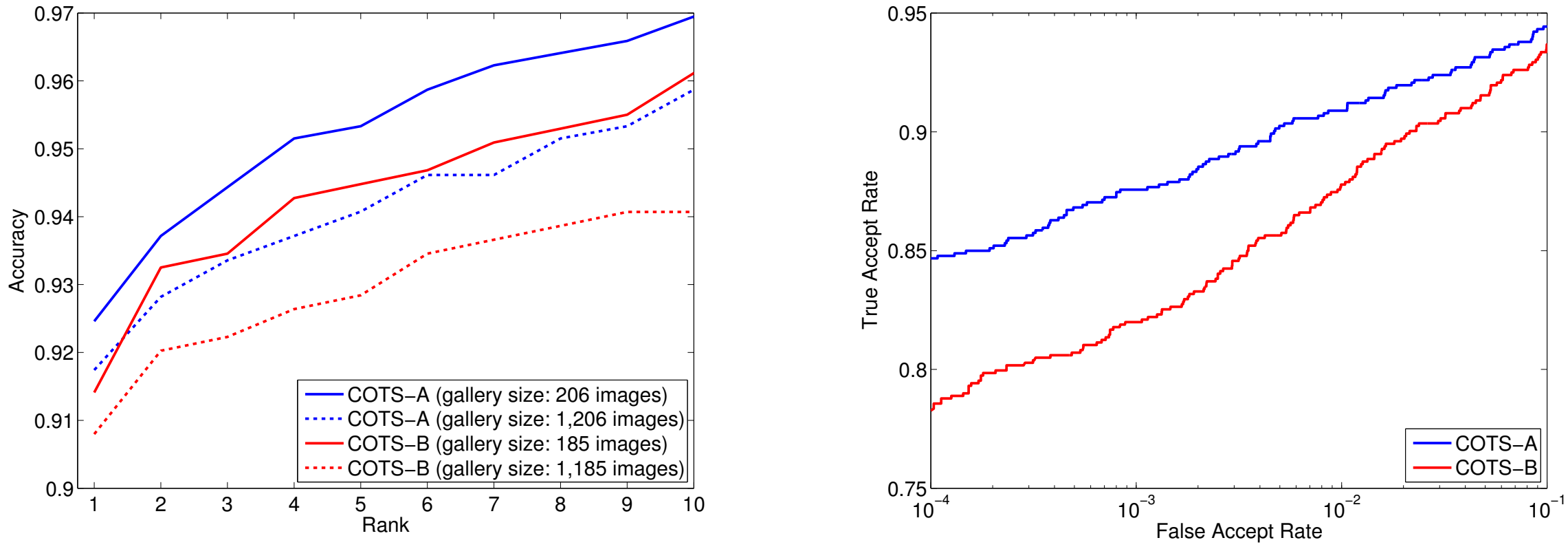

(a) (b)

*Figure 16: Results of face recognition experiments for (a) identification and (b) verification scenarios.*

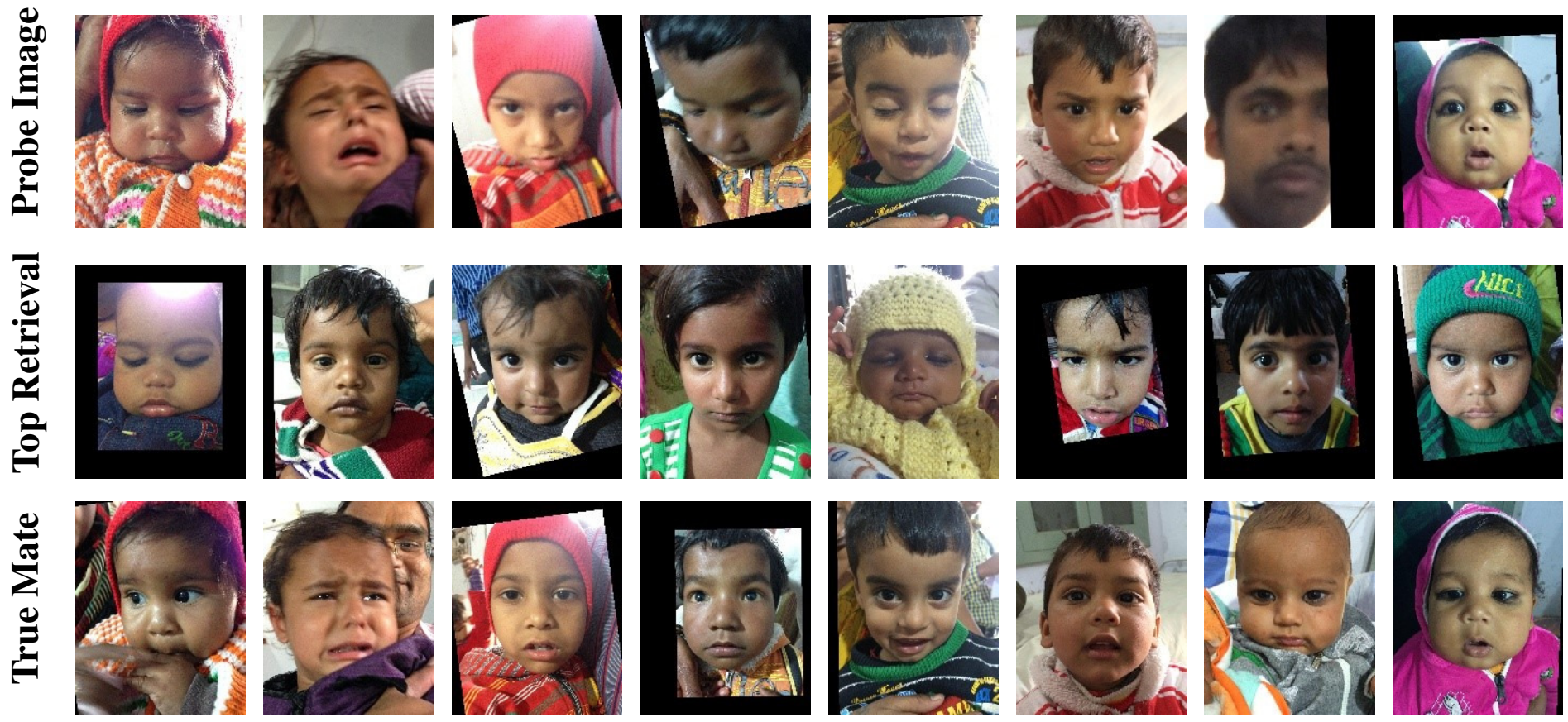

Rank-31 (a) | Rank-170 (b) | Rank-59 (c) | Rank-139 (d) | Rank-27 (e) | Rank-93 (f) | Rank-171 (g) | Rank-87 (h)

*Figure 17: Example face retrieval results. All examples shown are retrieval errors made by COTS-B against a gallery of 185 face images (21 out of 206 face images failed to enroll). Errors are due to variations in pose, illumination, and expression. In probe image (g), a background adult face was enrolled, rather than the foreground face of the child.*

### 3.2. Matching Face Images

During data collection at the hospital, 3-5 face images of each subject were captured in succession (manual shutter operation instead of video capture mode) over a time interval of approximately 60 seconds. Many of the acquired face images are quite unconstrained, particularly with respect to poor illumination conditions in the hospital room (see Figure 2(c)) and uncooperative subjects with variations in pose and expression. For face matching experiments, all face images were resized to 250×333, and face match scores were obtained from two COTS face matchers, denoted COTS-A and COTS-B. Figure 19 shows that poor illumination, motion blur, partial face images, and extreme facial pose caused failure to enroll (FTE) for a substantial number of face images (*i.e.* incorrect face and/or eye detections); COTS-A and COTS-B failed to enroll 42 and 123 of the 810 total face images, respectively.

To evaluate the performance of face recognition, we manually selected one face image per subject for the gallery. The remaining 604 face images of the 206 subjects were used as probes. We further extended the size of the gallery with 1,000 face images of approximately 200 young children of varying races (downloaded via Google image search). Face identification results in Figure 16(a) show that, as expected, extending the gallery decreases the identification accuracies of both COTS face matchers (by about 2%). The rank-1 accuracies of COTS-A and COTS-B are 91.74% and 90.80%, respectively. Figure 17 shows some examples where the true mate in the gallery was not retrieved at a low rank. Compared with fingerprint identification accuracies in Section 3.1.2, the performance of face identification is significantly lower, especially considering that the extended gallery for fingerprints is 32,768 fingerprints of 16,384 different fingers. Specifically, the rank-1 accuracy of the latent fingerprint SDK (COTS-L) using fusion of left and right thumbprint is 99.03% which is more than 7% higher than the rank-1 accuracy of the COTS-A face matcher.

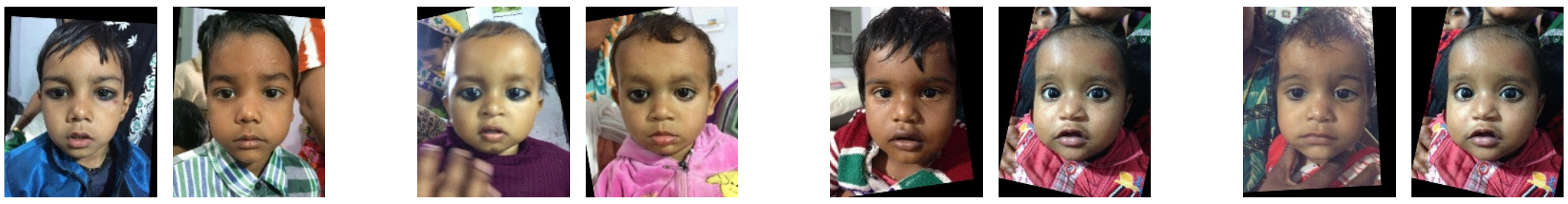

(a) (b) (c) (d)

*Figure 18: Examples of false accept errors made by COTS-A in face verification. The four examples shown are the impostor pairs of face images with the highest similarity as measured by COTS-A.*

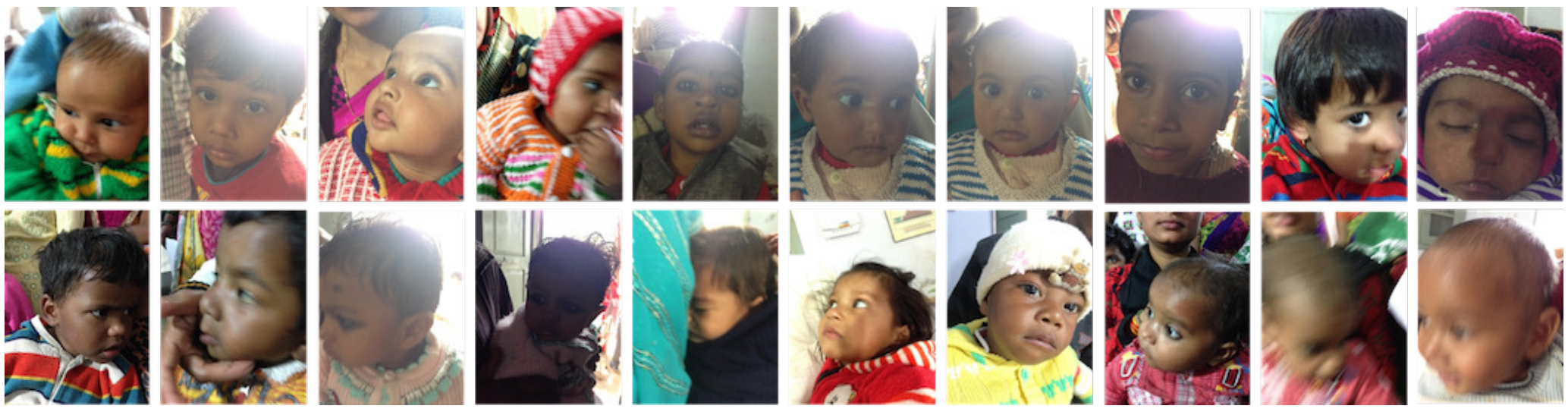

*Figure 19: Example face images which could not be enrolled by both the COTS face matchers (due to incorrect face and/or eye detections). This is due to unconstrained properties such as large non-frontal pose, poor illumination, and motion blur.*

Because the two COTS face matchers fail to enroll different numbers of face images, to better compare their capabilities, we evaluated face verification on all pairwise comparisons of face images that were enrolled by *both* COTS-A and COTS-B (669 out of the 810 total face images). Face verification results in Figure 16(b) show that the performance of COTS-A is significantly higher than the performance of COTS-B; COTS-A and COTS-B achieve TAR values of 87.57% and 81.99%, respectively, at 0.1% FAR. Some examples of false accepts made by COTS-A are shown in Figure 18.

Although these face matching results are less than satisfactory, especially considering that the face images represent same-day (in fact, same-minute) acquisitions, face images are still useful operationally. Healthcare workers (or other operators) can visually verify that the fingerprint matching results are correct using face images. For example, the top-K most similar face images could be shown to the healthcare worker for visual verification of the retrieval results (see Figure 20). Such roles for face images in the recognition of infants and toddlers will be investigated in later phases of this ongoing study.

## 4. Summary and Future Work

Current biometric data capture and recognition solutions cater primarily to adults (over 16 years of age). National ID programs, such as India's Aadhaar program, mandate capturing fingerprints and iris images of individuals who are 5 years of age or older. There is now a growing need for developing capabilities to recognize very young children (from newborn to 4 years old) based on biometrics. For biometric recognition to be successful, uniqueness and persistence properties of a biometric trait need to be satisfied for the population of interest. While these two fundamental tenets of biometrics have been investigated for the three primary biometric traits (fingerprints, face and iris) of the adult population, there is no comprehensive study conducted to investigate the persistence of biometric recognition (longitudinal study) for children. In this study, our goal is to investigate the persistence of fingerprint recognition for children in the age group of 0-4 years. For this purpose, we have completed the first of four phases in our effort to collect biometric data of children at Saran Ashram Hospital, Dayalbagh, Agra. In Phase I of the study, fingerprints and face images of 206 subjects were collected. We also established baseline biometric recognition performance for infants and toddlers using the captured fingerprints and face images. Based on our preliminary results, use of fingerprints for recognition of infants and toddlers appears promising. To comprehensively study the persistence of fingerprint recognition for children in the age group of 0-4 years, we plan to collect longitudinal data from the same subjects three more times over a one year period in September 2015 (Phase II), December 2015 (Phase III), and March 2016 (Phase IV).

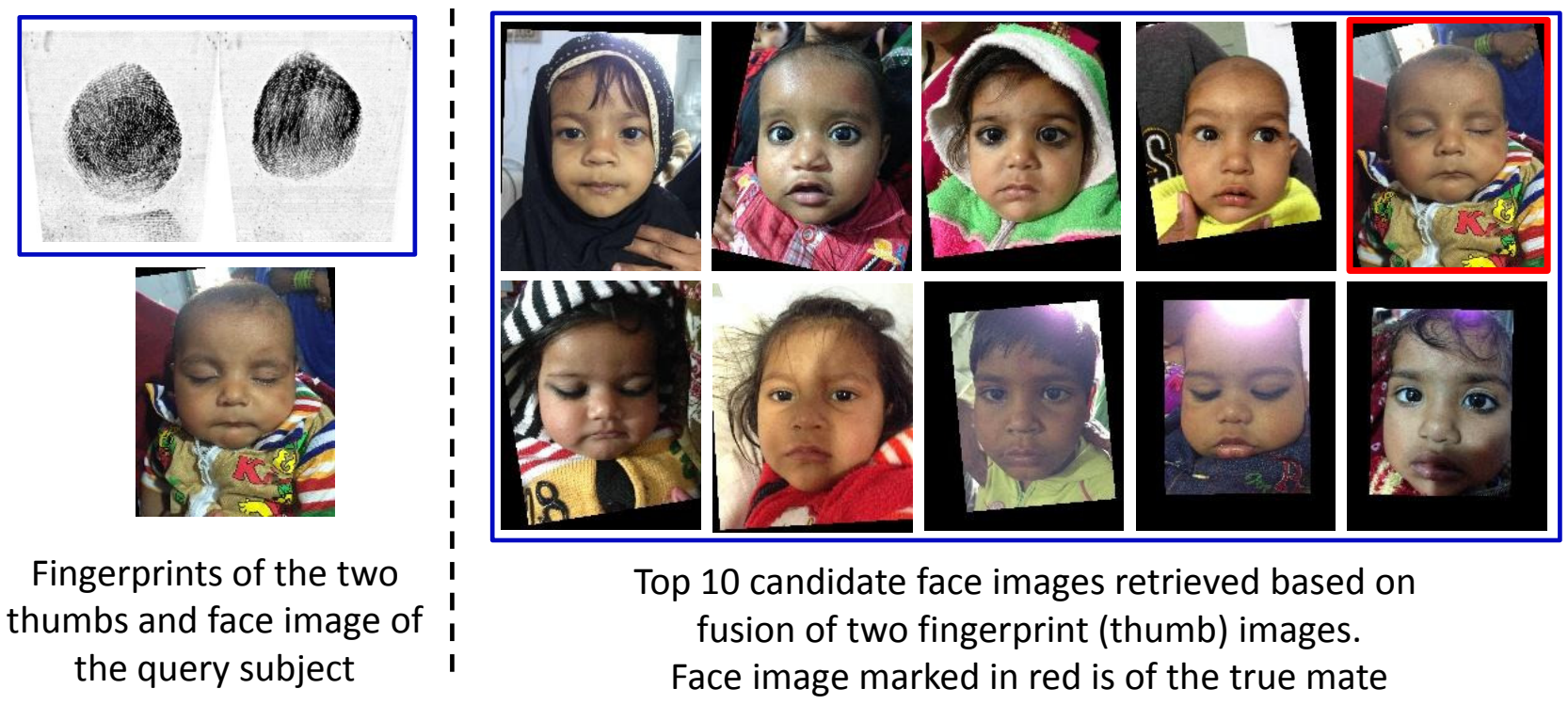

(a)

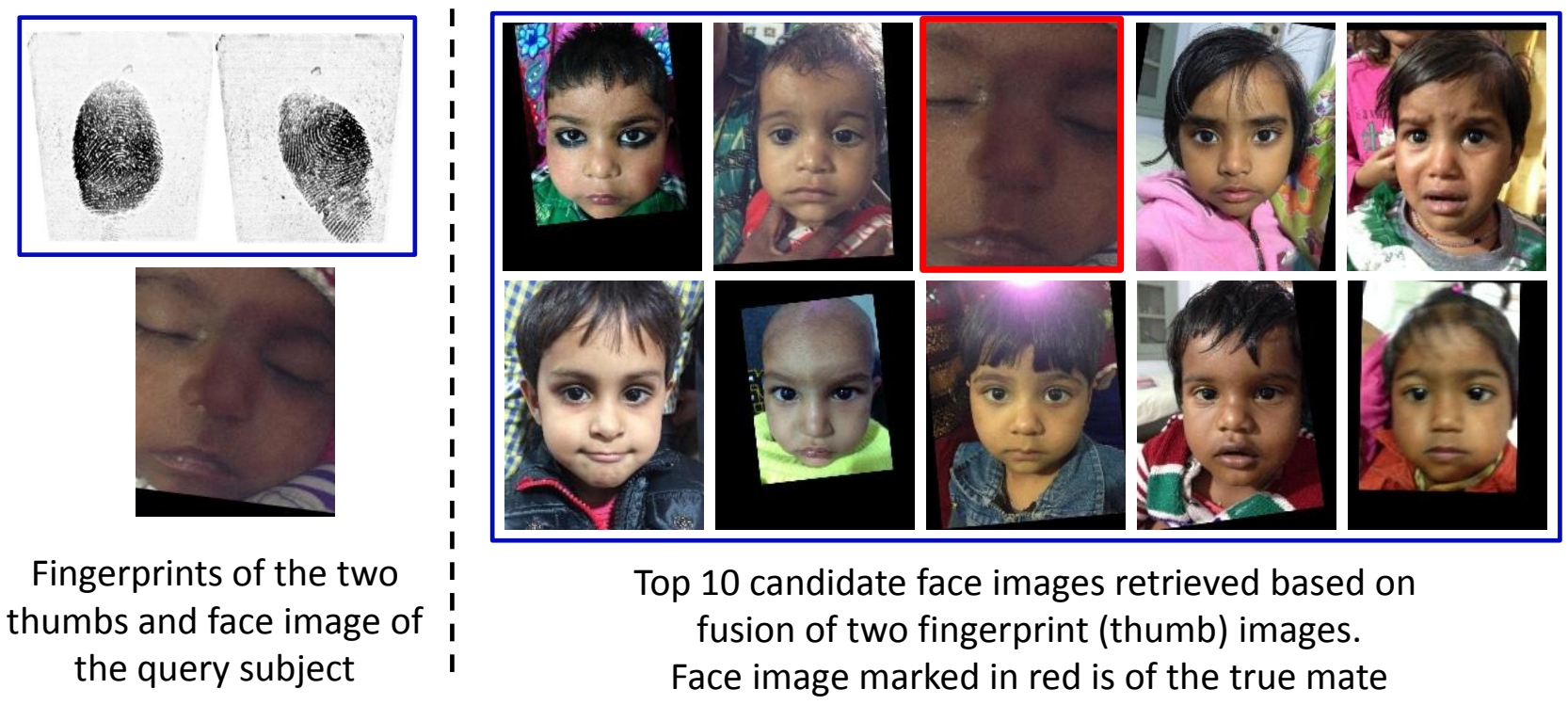

(b)

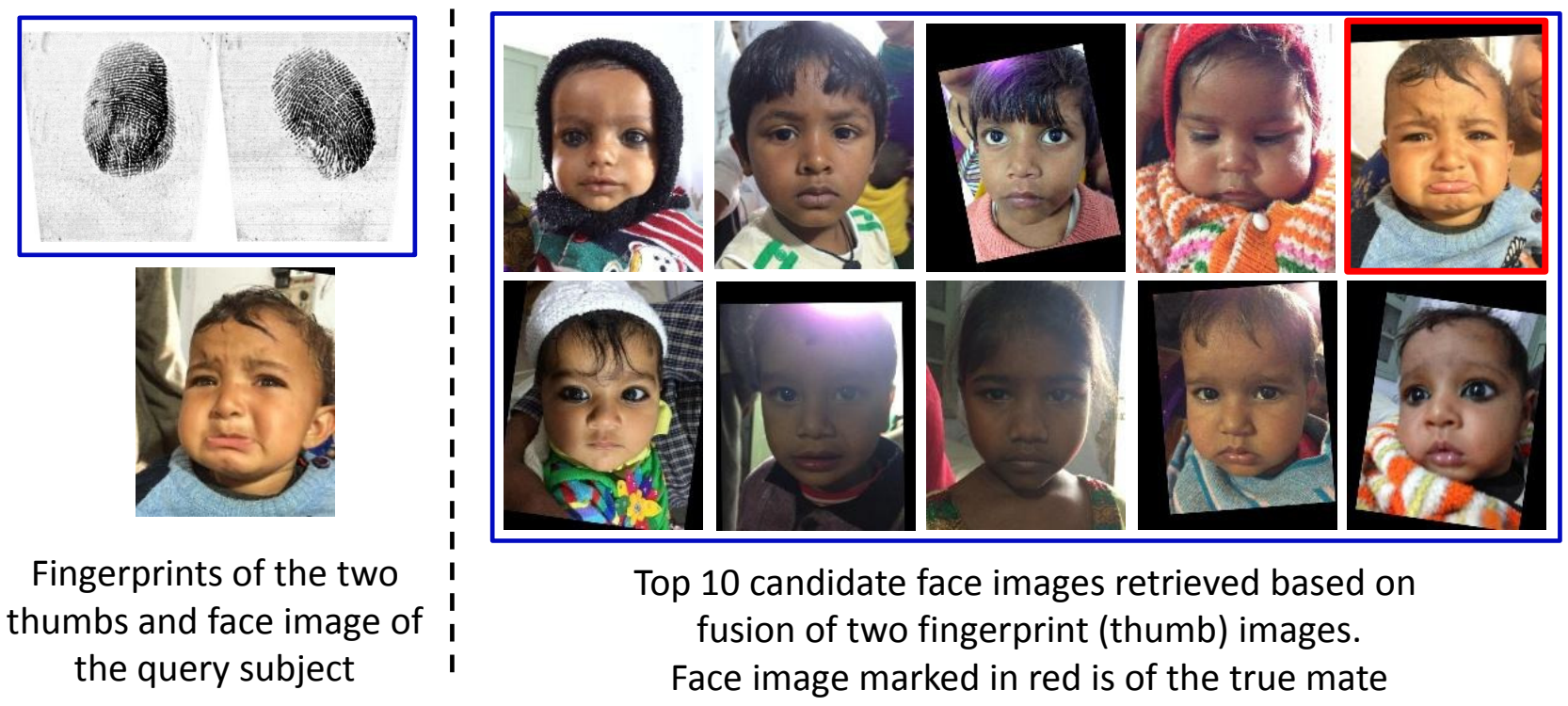

(c)

*Figure 20: Fingerprints and face image of three query subjects (shown on the left), and the face images of the top-K candidate matches using fingerprints shown to the operator for verifying the child's identity (shown on the right). The true mate is retrieved at ranks 5, 3, and 5, respectively for the three query subjects. The gallery (database) here consists of 206 subjects.*

## Acknowledgements

The authors would like to thank Ken Warman of the Bill and Melinda Gates Foundation and Mark Thomas of VaxTrac for their encouragement and support of this research. They would like to acknowledge Prof. Prem Kumar Kalra, Director, Dayalbagh Educational Institute, for his support of this study, and the staff and students of Dayalbagh Educational Institute for volunteering their time during data collection. The authors would also like to thank Dr. Vijai Kumar, Advisor, Medical Education and Healthcare Services, Dayalbagh and the staff of the Saran Ashram Hospital, Dayalbagh, who volunteered their time and provided logistical support for this study.